\documentclass[ASNA,twocolumn]{USG} 
\usepackage{anyfontsize} %

\usepackage{colortbl}   
\usepackage{xcolor}     
\usepackage{steinmetz}
\usepackage{graphicx} 

\graphicspath{{./images/}}

\articletype{ORIGINAL ARTICLE}%

\received{July 2026}
\revised{July 2026}

\begin{document}
\title{Towards Automated Initial Probe Placement in Transthoracic Teleultrasound Using Human Mesh and Skeleton Recovery}
\transtitle{Guidelines for Establishing a Cytometry Laboratory}
\subtranstitle{trans-subtitle}
\author[1]{Yu Chung Lee}[https://orcid.org/0009-0008-5297-4815]
\author[1]{David G. Black}[]
\author[1]{Ryan S. Yeung}[]
\author[1]{Septimiu E. Salcudean}[]

\authormark{Yu Chung Lee \textsc{et al.}}
\titlemark{Towards Automated Initial Probe Placement in Transthoracic Teleultrasound Using Human Mesh and Skeleton Recovery}

\address[1]{\orgdiv{Department of Electrical and Computer Engineering, }\orgname{The University of British Columbia, }%
\orgaddress{\state{Vancouver, British Columbia, }\country{Canada}}}

\corres{Yu Chung Lee (\email{lycpaul@ece.ubc.ca})}


\transkeywords{Teleultrasound | Mixed Reality | Patient Registration | Human Mesh Recovery | Probe Placement Guidance | FoCUS | POCUS}

\abstract[ABSTRACT]{Cardiac and lung ultrasound are technically demanding because operators must identify patient-specific intercostal acoustic windows and then navigate between standard views by adjusting probe position, rotation, and force across different imaging planes.
  These challenges are amplified in teleultrasound, where the examination proceeds without in-person expert assistance: once the probe is approximately positioned, the expert can navigate in ultrasound image space, but guiding the initial placement remotely remains difficult given the limited 3D perception of the patient.
  We present a framework for automating \textbf{P}atient registration and anatomy-informed \textbf{I}nitial \textbf{P}robe placement \textbf{G}uidance (PIPG) using RGB images obtained from a calibrated camera and a point cloud accumulated from depth images.
  The novice first captures the patient using the camera on a mixed reality (MR) head-mounted display (HMD), and an edge server then infers a patient-specific body-surface and skeleton model.
  By leveraging the patient's spatial and temporal consistency across multiview and point cloud data, we achieved robust, training-free human registration, verified in a healthcare setting.
  Using bony landmarks from the predicted skeleton, we estimate the intercostal region and project the guidance back onto the reconstructed body surface. To validate the framework, we rendered the reconstructed body mesh and the virtual probe pose guidance in the MR headset across multiple transthoracic echocardiography scan planes {\em in situ} and measured the quantitative placement error.
  Pilot experiments with five healthy volunteers suggest that the proposed probe placement prediction and MR guidance yield consistent initial placement, with a mean surface error of 15 mm, positional error to palpated anatomical landmarks of 41.0 mm, and torso orientation errors within 9°, acceptable for the teleultrasound setup.
  The codebase for the framework is available at: [RELEASED UPON PUBLICATION].
}

\abbr{5-FU, 5-fluorouracil; CFD, computational fluid dynamics; CH, channel; EFS, event-free survival; GBM, glioblastoma multiforme; OS, overall survival; PFS, progression-free
survival; SD, standard deviation.}



\copyright{This is an open access article under the terms of the \href{Creative Commons Attribution-NonCommercial}{Creative Commons Attribution-NonCommercial} License, which permits use, distribution, and reproduction in any medium, provided the
  original~work~is~properly cited and is not used for commercial purposes.
  \\[5pt]
©  2026 The Author(s)}


\maketitle


\section{Introduction}\label{sec:intro}

Point-of-care ultrasound (POCUS) offers a practical bedside workflow when comprehensive imaging is not immediately available, and focused cardiac ultrasound (FoCUS) targets a subset of views from transthoracic echocardiography (TTE) to support rapid assessment \cite{spencerFocusedCardiacUltrasound2013}.
Common FoCUS cardiac windows include parasternal long axis (PLAX), parasternal short axis (PSAX), apical four-chamber (A4C), subcostal four-chamber, and subcostal inferior vena cava (IVC) views \cite{uoftpiefocus2020}, while lung POCUS involves the left and right lungs' anterior, posterior, and costophrenic views \cite{uoftpiepocus2013}. Operators may reposition the patient between supine and left lateral decubitus to improve acoustic access, and frequently relocate the probe across intercostal spaces while adjusting angulation and rotation.

Teleoperated robotic or teleguided ultrasound systems have been proposed to improve access in remote areas and reduce the need for on-site expertise \cite{jiangRoboticUltrasoundImaging2023,marwickCurrentStatusPending2025a}.
They enable an expert to supervise a novice via model-mediated behaviors and to remotely teleoperate a robotic ultrasound system or provide remote visual guidance, thereby improving supervision and standardization \cite{jiangRoboticUltrasoundImaging2023,munirSurveyAutonomousRobotic2025,blackStabilityTransparencyMixed2024,yeungMeasurementPotentialFieldBased2025b}.
However, these systems often assume the bedside healthcare worker can already find a reasonable initial window, whereas in practice, the initial placement is a key bottleneck that significantly delays the expert’s ability to take over effectively \cite{roshanFindingInitialProbe2025}.
It is also difficult to initially locate the probe in teleoperated ultrasound on the patient via teleconferencing, given the limited patient view and the lack of 3D vision.
Recent work has begun to automate parts of the probe initialization problem.
The search-based approach \cite{laurentEchoRobotSemiAutonomousCardiac2025} uses a predefined search pattern but is inherently view-specific.
Pure point cloud-based approaches \cite{roshanFindingInitialProbe2025,nardiAnatomyAwareSharedControl2026} can infer the chest and patient surface but are limited to the supine pose and generalize poorly to other postures due to their inability to handle complex postures and torso shape ambiguity.
A recent SAM 3D Body-based approach \cite{yangSAM3DBody} pretrained on a large-scale dataset for Human Mesh Recovery (HMR) tasks achieves outstanding performance on general scenes, but clinical applications remain challenging due to the rarity of in-bed postures in the dataset and limited clinical data due to privacy concerns. Some researchers attempt to leverage the temporal consistency of patient posture \cite{tu2026patient4dtemporallyconsistentpatient}, but it remains unclear whether absolute depth information can be recovered, a critical requirement for accurate ultrasound probe placement on the patient's surface.
Thus, automating initial POCUS probe placement remains an unsolved problem.

To address this unmet need and streamline the setup process, we propose an mixed reality-based approach that (i) reconstructs and registers a patient-specific body model from RGB imagery and point cloud captured by an HMD, (ii) estimates anatomy-referenced landmarks and generates initial probe pose guidance for multiple transthoracic scan planes, specifically for POCUS cardiac and lung ultrasound, and (iii) visualizes guidance {\em in situ} to support novice setup for most forms of teleultrasound. To our knowledge, this is the first demonstration of MR guidance that leverages a human skeleton model derived from mesh recovery to estimate probe placement across diverse scan planes.
In summary, our approach comprises and integrates the following:

(i) \textbf{Huuman Mesh Recovery}, which, with parametric body models, enables estimation of patient pose and shape from RGB images. We build on recent advances in SAM-based mesh recovery \cite{yangSAM3DBody,gaoSAMBody4DTrainingFree4D2025} and parametric representations such as SMPL/SMPL-X \cite{loperSMPLSkinnedMultiPersona,pavlakosExpressiveBodyCapture2019}.
We further infer a biomechanical skeletal structure and localize bony landmarks using the SKEL model, which re-rigs SMPL with an anatomically accurate skeleton \cite{dakriPredicting3DBone2024a}.

(ii) \textbf{Robust HMR with multiview RGB and point cloud}, our algorithm uses multiview RGB predictions as 3D priors and remaps them into a unified coordinate system to leverage spatial consistency. We then employ iterative closest point (ICP) \cite{ICP121791} with depth constraints to align the mesh with the point cloud. We further perform body-part segmentation using the nearest neighbors of mesh vertices and use orphan vertices as unseen bounds to obtain a robust SMPL-X body fitting aligned with point cloud measurements.

(iii) \textbf{Anatomy-Aware Guidance}, or body-surface constrained guidance, which is relevant when translating predicted poses into actionable contact points on the torso \cite{nardiAnatomyAwareSharedControl2026,kellerSkinSkeletonBiomechanically2023a}. Our work uses reconstructed surface and skeletal landmarks to generate probe placement cues that lie on the patient's surface and remain consistent across posture changes.

(iv) \textbf{Mixed Reality}, remote guidance methodology that can tightly couple an expert and a novice via {\em in situ} visual cues, enabling hand-over-hand guidance to improve visualization and interaction in teleultrasound \cite{blackRoboticHumanTeleoperation2025,blackStabilityTransparencyMixed2024,yeungMeasurementPotentialFieldBased2025b}.

\begin{figure*}[htb]
  \centerline{\includegraphics[width=0.9\linewidth]{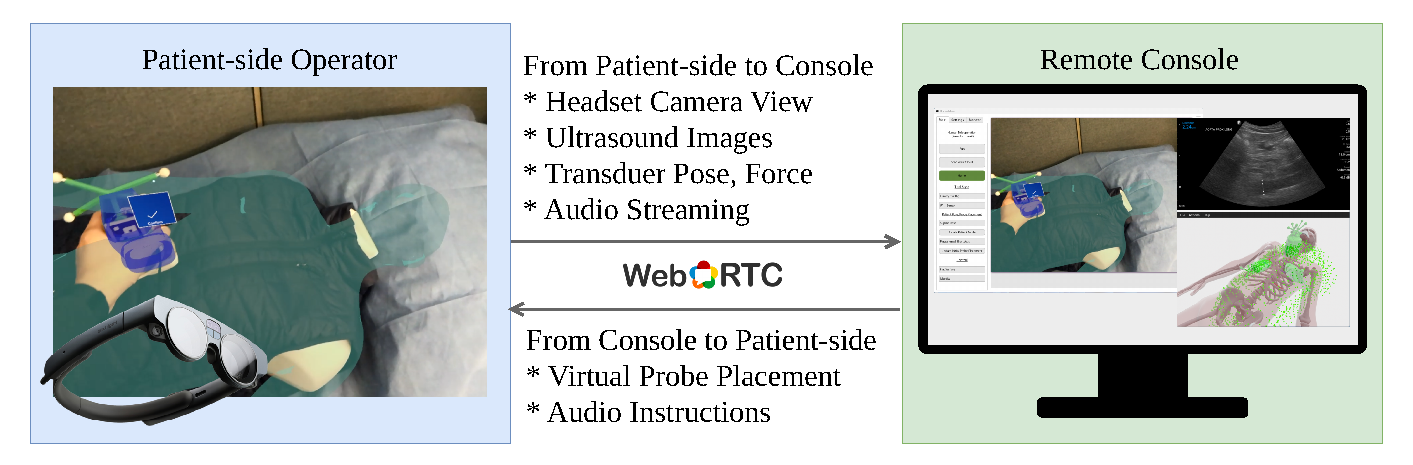}}
  \caption{Teleultrasound's patient-side and remote console with bidirectional streaming and MR guidance overlays via the WebRTC protocol. The remote console end could monitor the examination from the headset's point-of-view camera and ultrasound video.}
  \label{fig:system}
\end{figure*}

\section{Methods}\label{sec:methods}

\subsection{Problem Formulation}
Given a patient lying on an exam bed in one of several clinically relevant postures (e.g., supine or left lateral decubitus), our objective is to generate an initial POCUS probe placement pose for a set of target transthoracic scan planes.
A patient-side novice captures multiple RGB images and an accumulated point cloud with a calibrated RGB-D camera, and the system returns: (a) a registered patient body surface mesh and skeleton model in world coordinates, and (b) a probe pose for each target scan plane, projected on the recovered patient surface.
These outputs are visualized as {\em in situ} holographic overlays to support initial setup before (or while) the remote expert or the robotic system provides fine guidance.

\subsection{System Overview}

\begin{figure*}[htb]
  \centering
  \includegraphics[width=0.85\linewidth]{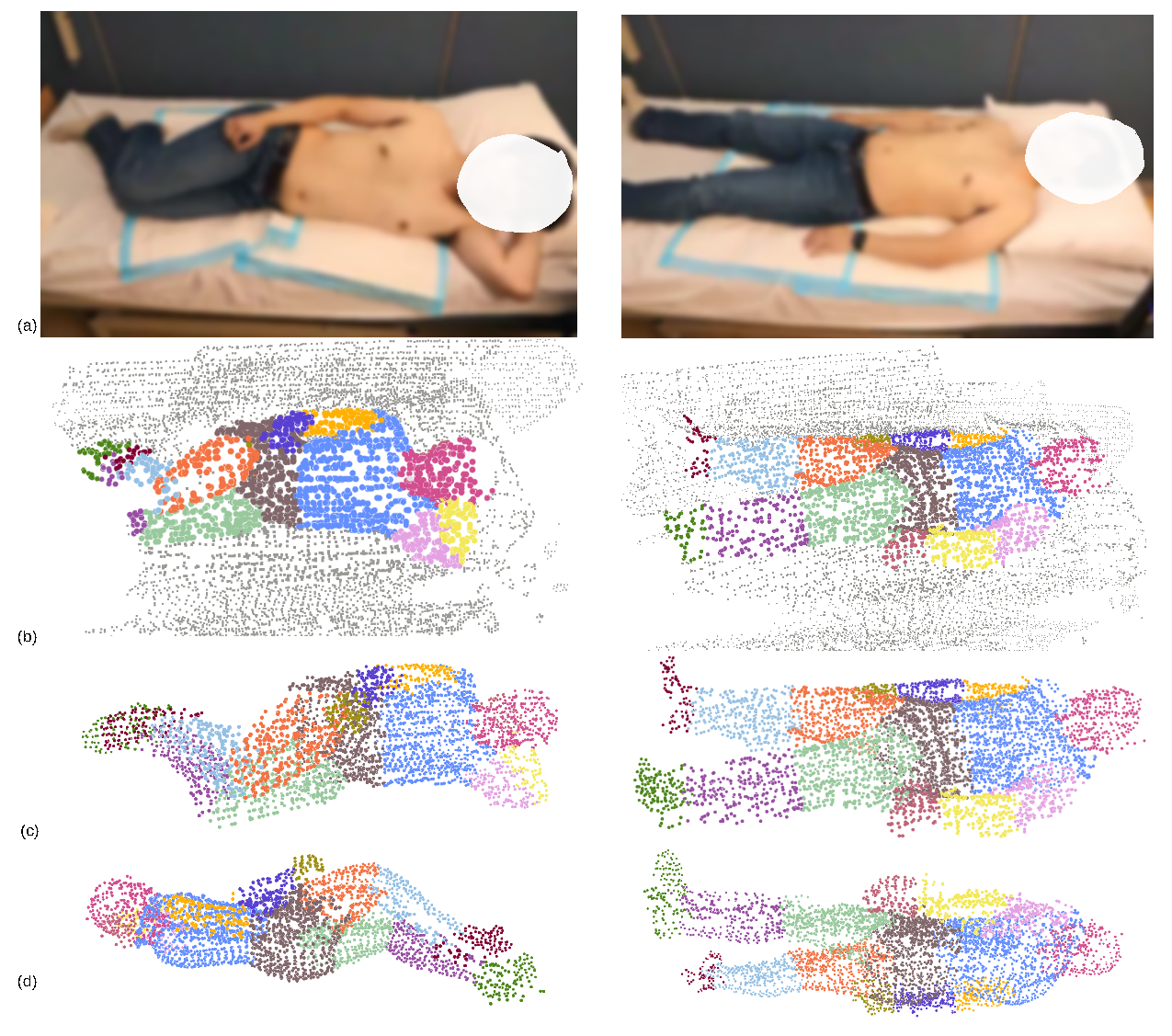}
  \caption{Body-part point cloud segmentation via majority-vote label transfer from nearby MHR mesh vertices. Left column: left lateral decubitus; right column: supine. From top to bottom rows: (a) RGB image captured by MR headset, (b) recorded point cloud with body part segmentation, (c) segmented point cloud $\mathcal{P}_{\text{bparts}}$ with orphan vertices $\mathcal{V}_{\text{orphans}}$, (d) back view of the combined points.}
  \label{fig:segment}
\end{figure*}

Figure~\ref{fig:system} illustrates the system architecture.
On the patient side, an MR headset (Magic Leap 2) provides first-person view (FPV) camera streams and ultrasound images to the remote console using a real-time communication stack.
The headset produces holographic overlays of the virtual probe as described in \cite{blackRoboticHumanTeleoperation2025}.
We attached ArUco markers on the ultrasound probe for pose measurements and tracking \cite{garrido-juradoAutomaticGenerationDetection2014}.
The remote console runs on a standard office desktop and transfers registration metadata to the edge server. The guidance prediction module, including model inference and optimization, runs on an edge server equipped with an NVIDIA RTX 3090 Ti.
The overall framework is implemented in PyTorch for machine learning and optimization components, with ZeroMQ as middleware used to package the inference stacks for deployment and monitoring.
Our overall framework, which uses body-part segmentation (Fig.~\ref{fig:segment}), is described in Figure~\ref{fig:pipeline}.

\subsection{Automatic Patient Registration}
We estimate a patient-specific body model from RGB images captured across multiple views and align it with the MR headset coordinate frame.

\textbf{Coordinate system.} Let $\{W\}$ denote the HMD world frame, with the HMD FPV camera frame $\{C\}$,  the reconstructed body/skeleton canonical frame $\{B\}$, and $\{P_i\}$ denote the probe pose of the $i$th view. The HMD provides the camera pose homogeneous transformation $^{W}\mathbf{T}_{C}$. A human mesh recovery model predicts a body representation and yields translation and orientation $^{C}\mathbf{T}_{B}$ with respect to the local camera frame. We obtain the overall patient transformation with $^{W}\mathbf{T}_{B} = \, ^{W}\mathbf{T}_{C} \; ^{C}\mathbf{T}_{B}$ and the desired probe transformation output by the prediction model is $^w\mathbf{T}_P$. The point cloud $\mathcal{P}$ is constructed by aggregating depth measurements from one or more depth frames acquired with the calibrated depth camera on the HMD.
For each observation, the corresponding depth points are transformed into the HMD world coordinate system by applying the camera pose $^{W}\mathbf{T}_{C}$.

\textbf{Multiview SAM 3D Body as initial prediction.} We first use SAM 3D Body \cite{yangSAM3DBody} to recover 3D human meshes represented in Momentum Human Rig (MHR) parameters \cite{fergusonMHRMomentumHuman2025}.
This stage takes a series of eight RGB images with coherent body shape estimates $\beta \in \mathbb{R}^{45}$ and applies a Kalman filter to the body's joint pose estimates to leverage temporal and shape consistency \cite{yang2026fastsam3dbody}.
Camera intrinsics are obtained from the HMD and provided to the model, rather than using the default field-of-view estimator. The recovered MHR meshes $\mathcal{M}_{\text{mhr}}$ are then mapped to the HMD world frame for further processing.

\textbf{Body-part point cloud segmentation.}
From our pilot study, we noticed a systematic bias in the depth estimation using SAM 3D Body with images acquired by the MagicLeap camera. We thus used ICP constrained to adjust along the camera optical axis (z-axis outward along the camera pose), to improve depth perception of the MHR meshes, with the point-to-plane objective function to obtain the optimal transform of each mesh $T_\delta (m_i) = m_i + \delta \hat{d}$ formulated as
$$
\operatorname*{min}_{\delta \in [\delta_{\min}, \delta_{\max}]} \sum_{i \in \mathcal{M}} w_i \Big[ \mathbf{n}_i^{\top} \big( \mathbf{m}_i + \delta \hat{\mathbf{d}} - \mathbf{q}_i \big) \Big]^2 + \lambda (\delta - \delta_0)^2
$$
across a set of meshes $\mathcal{M}_{\text{mhr}}$, with a scalar depth offset $\delta \in [\delta_{\min}, \delta_{\max}]$, $\hat{d}$ is the camera optical axis direction extracted from the camera extrinsic, and the estimated target normal $n_i$ of the nearest point $q_i$. The weights $w_i$ are the robust iteratively re-weighted least-squares weights from the Huber loss, and $\lambda$ is the $L_2$ regularization weight away from the depth prior $\delta_0$. The closed-form update per iteration:

\[
\delta^{*} = \operatorname{clip}\left(
-\frac{\sum_i w_i c_i e_i - \lambda \delta_0}{\sum_i w_i c_i^2 + \lambda}, \delta_{\min},\delta_{\max} \right)
\]
\[
c_i = n_i^\top \hat{d} \quad e_i = n_i^\top (m_i - q_i)
\]

The ICP registered meshes set $m_i\in\mathcal{M}_{\text{reg}}$ obtained from the closed-form update, iteratively by recomputing the next nearest neighbors $q_i^{(k)}$ and new correspondences $c_i^{(k)}$ at each cycle $k$ to update the meshes tranformation $m_i^{(k)} = m_i + \delta^{(k)}\hat{d}$.

We use the body part vertices definition from \cite{takmaz20233dsegmentationhumanspoint} to compute the centroid of each body part $c_k = \frac{1}{N}\sum^{N_k}_{i=1}\mathbf{p}_i$ as the initial global body transformation in the later fitting stage. To further incorporate the point cloud observation into parametric model fitting, we assign part labels to the point cloud as in \cite{lascheit2025robusthumanregistrationbody} by a majority-vote scheme over its neighboring MHR vertices from $\mathcal{M}_{\text{reg}}$ with inverse distance weight $\alpha_j$ and Kronecker delta $\mathbb{1}[ \cdot ]$:
$$
s_i = \operatorname*{arg\,max}_{k \in \{1, \dots, K\}} \sum_{j=1}^{n} \alpha_j \, \mathbb{1}\big[\text{label}(\mathbf{v}_j) = k\big]
$$
MHR vertices that are not assigned to any points $\mathcal{V}_{\text{orphans}}$ will be treated as unobserved points and used in the fitting stage with a discount factor. The segmented points $\mathcal{P}_{\text{bparts}}$ with semantic labels will be used in the body-part matching loss during the fitting process, as described in the next section.

\subsection{Human Mesh and Skeleton Recovery}

\textbf{SMPL-X Body Model Fittings.}
The SMPL-X model, defined as $M_{\mathrm{SMPLX}}(\beta, \theta, \mathbf{t})$ with shape parameters $\beta \in \mathbb{R}^{10}$, body pose parameters $\theta \in \mathbb{R}^{22 \times 3}$, and global translation $\mathbf{t} \in \mathbb{R}^{3}$, is used to render the body mesh in the HMD using the SMPLX-Unity library provided by \cite{pavlakosExpressiveBodyCapture2019}. We follow \cite{lascheit2025robusthumanregistrationbody} to incorporate losses into the objective function. In addition, we introduce a segmented body part point cloud and unobserved orphan vertices obtained from the SAM 3D Body prior. The overall objective function, with hyperparameters $\lambda_{\text{term}}$ corresponding to the weighting of different sub-objectives, is given by
$$
\mathcal{L} = \lambda_{\text{bparts}} \mathcal{L}_{\text{bparts}} + \lambda_{\text{orphans}} \mathcal{L}_{\text{orphans}} + \lambda_{\text{vposer}} \mathcal{L}_{\text{vposer}} + \lambda_{\text{shape}} \mathcal{L}_{\text{shape}}
$$
where the cost components are as follows:\\
\textit{Body-part matching} $\mathcal{L}_{\text{bparts}}$, measuring the model surface alignment with the segmented point cloud $\mathcal{P}_{\text{bparts}}$. Rather than a symmetric two-way distance, it uses a one-directional nearest-vertex matching per body part, wrapped in a Huber loss for robustness to outliers.
$$
\mathcal{L}_{\text{bparts}} = \sum_{k=1}^{K} \sum_{i=1}^{N_k} \min_{\mathbf{v} \in \mathcal{V}_k} \mathcal{L}_{\text{Huber}}(\mathbf{p}_i - \mathbf{v})
$$

\textit{Orphans vertices} $\mathcal{L}_{\text{orphans}}$, is similar to $\mathcal{L}_{\text{bparts}}$, but uses the unmatched vertices $\mathcal{V}_{orphans}$ to penalize mesh surface fitting on the unobserved region with a discount factor. 
It generally provides a reasonable bound on the body shape and avoids overfitting to the point cloud, especially when the captured point cloud is limited by the viewing angle and the in-bed posture.

\textit{VPoser regularization} $\mathcal{L}_{\text{vposer}}$ \cite{pavlakos2019expressivebodycapture3d}, a pretrained human pose prior constrained on plausible joint articulation, discouraging unrealistic body proportions.
$$
\mathcal{L}_{\text{vposer}} = \| \boldsymbol{\theta} - \boldsymbol{\theta}_0 \|_2^2
$$

\textit{Body shape regularization} $\mathcal{L}_{\text{shape}}$, penalizing on extreme, unrealistic human shape.
$$
\mathcal{L}_{\text{shape}} = \| \boldsymbol{\beta} \|_2^2
$$

The recovered SMPL-X model parameters are then sent to the MR headset to render the patient's surface for verification and to generate the skeleton model.

\textbf{Skeleton model conversion.}
The fitted SMPL-X model is used to obtain the SKEL model \cite{kellerSkinSkeletonBiomechanically2023a} for predicting the rib cage pose and structure.
Given that the SKEL anatomical joint regressor is tightly constrained to the SMPL vertex set $\mathcal{V}_{\text{SMPL}} \in \mathbb{R}^{6890 \times 3}$ \cite{loperSMPLSkinnedMultiPersona} and the associated shape parameters $\beta \in \mathbb{R}^{10}$, we employed the barycentric model–vertex conversion utility together with the analytical parameter conversion solver provided by the SOMA library \cite{soma2026}. This procedure supplies shape parameters as an informed initialization (warm start) and generates SMPL-like vertices for the alignment module.
To facilitate the optimization loop, we partially mask vertices and joints, retaining only the torso region when computing the fitting loss. This conversion provides a consistent set of joint locations and bony landmarks for generating downstream probe placements.
Both the optimized SKEL aligner and the SOMA-based model conversion tools will be available in our released source code.

\begin{figure*}[htb]
  \centering
  \includegraphics[width=\linewidth]{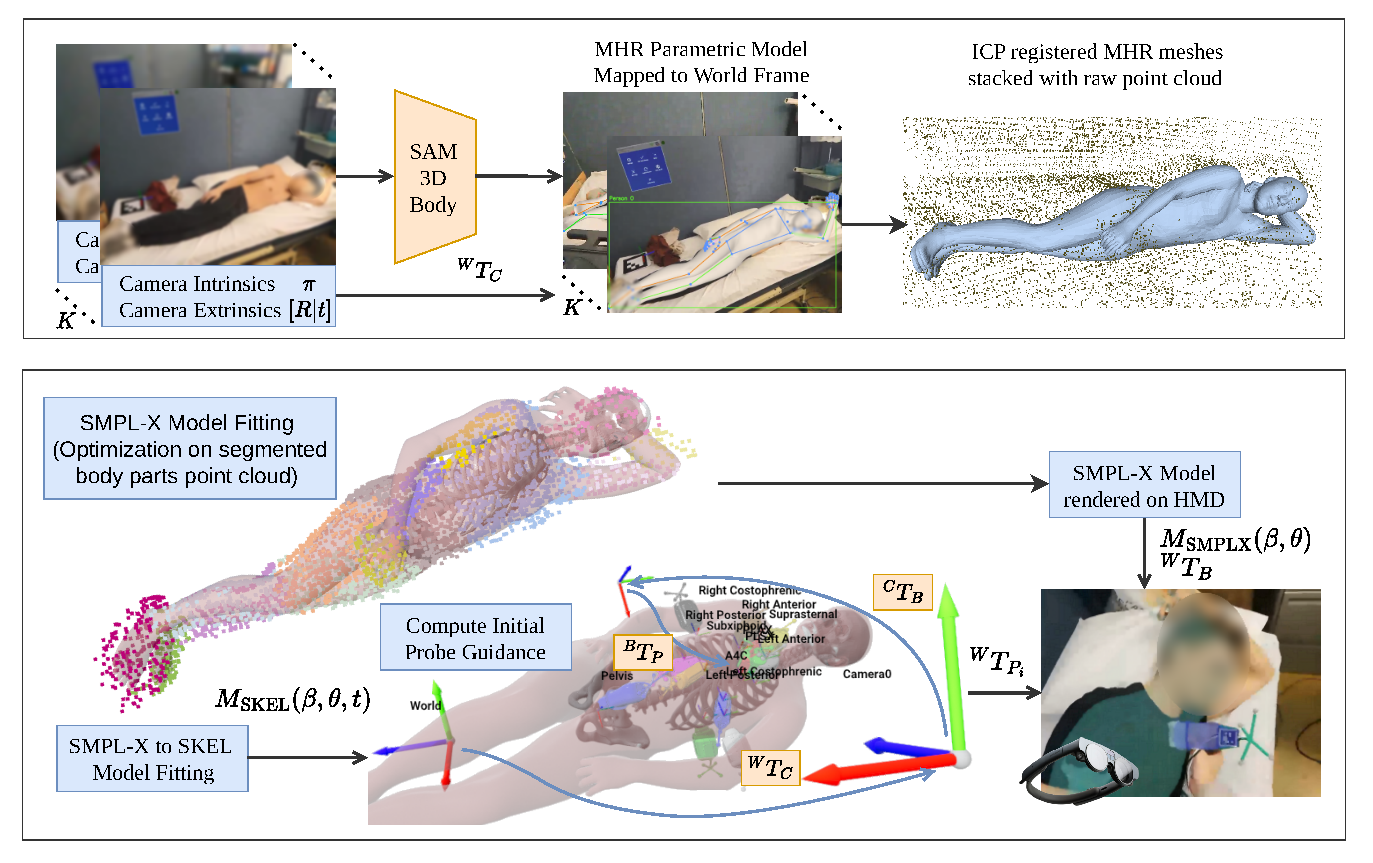}
  \caption{Overall framework of the proposed Patient Registration and Initial Probe Placement Guidance (PIPG) pipeline. Given a batch of $K$ RGB images and an associated point cloud, the system exploits the patient’s spatial and temporal consistency to estimate the patient-specific anatomy and register it to the system coordinate frame for real-time guidance generation.
  The first processing block illustrates the RGB image stream, from which MHR predictions are obtained using SAM 3D Body. These predictions are subsequently registered to the point cloud via Iterative Closest Point (ICP). The second processing block depicts the reconstruction of the SMPL-X model from the segmented point cloud, partitioned by anatomical regions as detailed in Section~\ref{sec:methods}. The recovered SMPL-X model is then employed to compute the initial probe placement guidance, while both are rendered in the MR headset.}
  \label{fig:pipeline}
\end{figure*}

\textbf{Runtime optimization}
The current implementation relies on optimization-based fitting and parametric model conversion, which can dominate runtime. To improve efficiency, we leverage a recent accelerated SAM 3D Body inference framework \cite{yang2026fastsam3dbody} and an analytical model conversion tool from SOMA-X \cite{soma2026}. We further accelerate the original SKEL fitting by adding caching and reducing per-iteration overhead. Combined with restricting the fitting loss to the torso, these optimizations reduce the end-to-end runtime of the pipeline in Figure~\ref{fig:pipeline} to within 20\,s on our system, compared with the over 2\,min runtime reported in related work \cite{nardiAnatomyAwareSharedControl2026}. Because registration is a one-off process during the initial placement planning phase, real-time inference is unnecessary, and the additional processing time introduces only a limited procedural delay.

\subsection{Probe Placement Guidance Generation}
For each target scan plane, we predefine an anatomy-referenced rule that maps skeleton/bony landmarks to an initial probe pose.
The transformation from patient body frame to each probe view  $^{B}\mathbf{T}_{P_i}$  can be deduced by the following cues:
(i) a finite set of bony landmarks (e.g., sternum and adjacent rib/intercostal references),
(ii) an offset model along the thoracic surface normal, and
(iii) a desired probe orientation relative to anatomical axes (e.g., along the pelvis or thoracic axis).

\textbf{Anatomy-Aware Probe Pose.}
Let $\mathbf{p}_{\ell}$ be a predicted landmark in the body frame, and let $\hat{\mathbf{n}}(\mathbf{p})$ denote the outward normal with respect to the thoracic axis.
To ensure that the virtual guidance lies on the patient surface, we compute the final contact point by projecting the landmark along the outward normal until it intersects the reconstructed torso surface, and then place the probe at the intersection point with a small inward offset to indicate nominal contact pressure. We further orient each projected probe pose according to POCUS guidelines in \cite{uoftpiefocus2020} relative to the body longitudinal axis.

\section{Experiments}\label{sec:experiments}

\begin{figure*}[hbt]
  \centering
  \subfloat[(a) Experimental setup]{%
    \includegraphics[width=0.35\linewidth]{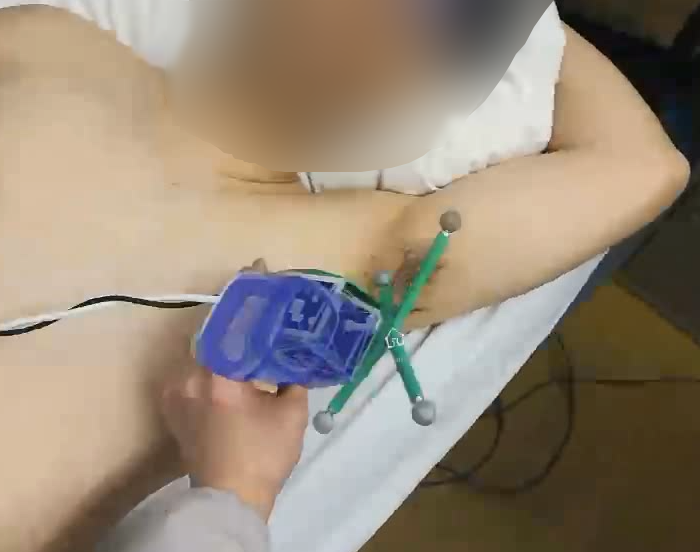}%
  }\hfill
  \subfloat[(b) Visualized recorded probe pose]{%
    \includegraphics[width=0.35\linewidth]{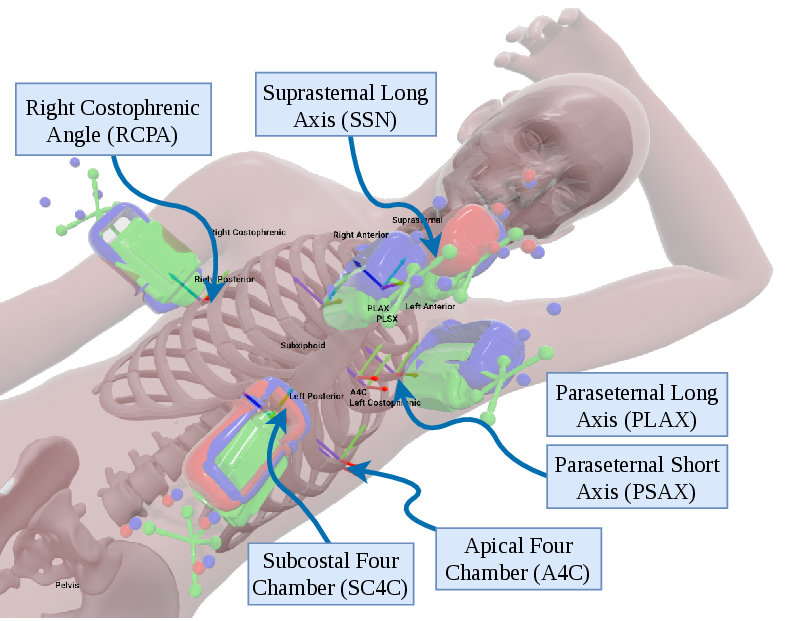}%
  }\hfill
  \subfloat[(c) Initial A4C views obtained by locating PMI manually]{%
    \includegraphics[width=0.2\linewidth]{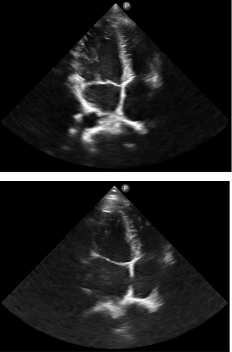}%
  }
  \caption{(a) First-person perspective from the MR headset during probe tracking and guidance. (b) The predicted probe pose is visualized in green, the guided pose in blue, and the ground-truth pose in red, corresponding to the upper and lower sternum and the point of maximal impulse (PMI). (c) The PMI is initially localized manually with the patient in the supine position and iteratively adjusted until the relevant anatomical structures can be approximately identified. Note that additional fine-tuning is typically required to achieve the final optimal imaging view.}
  \label{fig:pose_results}
\end{figure*}

We conducted pilot experiments with healthy volunteers in two postures: supine and left lateral decubitus.
The novice user wore an MR headset (Magic Leap 2) and captured the point cloud using the headset's depth camera, along with eight RGB images of the patient, each with an associated camera pose and parameters.
We then generated and visualized initial probe guidance $p_{\text{pred}}$ for selected POCUS cardiac and lung windows in each posture.
The novice user was asked to follow the visual guidance, maneuvering the probe to align with the suggested pose while maintaining contact with the patient’s body. We recorded the resulting probe pose as the \emph{guided pose} $p_{\text{guided}}$.
To obtain a ground-truth reference, we manually positioned the probe at the upper and lower sternal landmarks and the point of maximum impulse (PMI). The recorded pose is marked as the \emph{ground-truth anatomical pose} $p_{\text{gt}}$ because these landmarks are palpable and exhibit the least ambiguity.
We recruited five healthy volunteers as patients (five male, height $180 \pm 7$\, cm, weight $75 \pm 7$\, kg, age $25.5 \pm 3.5$ years), three of whom also served as novice users. We further analyzed the results based on the following three comparison metrics.

\textbf{a. Depth and surface reconstruction error ${e}_{depth}$.}
The novice user was instructed to align a physical probe with the predicted pose $p_{\text{pred}}$ and project it toward the patient's body until it contacted the skin with slight pressure, which was then marked as $p_{\text{guided}}$.
We measure the discrepancy between the MR-guided contact point and the predicted probe position, thereby capturing offsets in surface mesh recovery and depth estimation. Based on the accessibility and representativeness of each view, we measured the suprasternal long-axis (PLAX) and short-axis (PSAX), subcostal four-chamber (SC4C), right and left anterior (R/L-ANT), and costophrenic angle (R/L-CPA) views in the supine posture to cover various thoracic regions. In left lateral decubitus, the apical four-chamber and left costophrenic views were omitted because they were inaccessible in our setup.

\textbf{b. Anatomy alignment error ${e}_{anatomy}$.}
We selected identifiable landmarks (both visually and by palpation) and then used the tracked probe to record their positions as the ground-truth reference $p_{gt}$. 
Specifically, the upper and lower sternum and the point of maximal impulse (PMI) were chosen as the proximal starting points for SSN, SC4C, and A4C, respectively. 
We measure how well the predicted initial probe placement $p_{\text{pred}}$ aligns with the ground-truth reference.
When locating the upper and lower sternum, the operator orients the probe normal to the torso surface and aligns the image plane to the longitudinal axis. The recorded probe rotation is used as the ground-truth torso rotation and is compared with the predicted rib cage orientation.
We follow related cardiac exam instructions \cite{pocus101cardiac2020} to locate the PMI and use a Philips Lumify S4-1 phased array transducer to conduct a scan until we can roughly identify the mitral and tricuspid valve, left/right ventricle, and atrium. We do not attempt to obtain the best standard view, as the predicted pose is intended only for initial placement.

\textbf{c. Task-level utility error ${e}_{utility}$.}
We compare the novice user’s final probe position under MR guidance $p_{guided}$, with the ground-truth target location $p_{gt}$, to evaluate whether the guidance facilitates convergence toward the intended anatomical target.
The positional error quantifies the spatial discrepancy between the guided final probe placement and the corresponding reference anatomical landmark, namely the upper sternum, lower sternum, or PMI. This metric reflects the utility of the initial probe placement guidance in localizing the SSN, SC4C, and A4C imaging views, respectively.

\textbf{Positional and orientational error metrics.}
For each probe pose comparison, we report the positional error as Euclidean distance. The orientation error is reported as tilt error $e_{\text{tilt}}$, and the spin error $e_{\text{spin}}$.

Defining the probe normal of probe pose $p_\text{probe}$ as $\mathbf{n}_{probe} = R_{probe}\hat{\mathbf{z}}$, and the view plane normal denoted as $\mathbf{x}_{probe} = R_{probe}\hat{\mathbf{x}}$, the spin error is computed by projecting the view plane normal to the thoracic reference plane $\pi_{thorax}$ with normal ${\mathbf{n}}_{thorax} = R_{thorax}\hat{\mathbf{z}}$, where $R_{thorax}$ obtained by the mesh recovery.
We then compute all errors of two probe pose pair as $e_\text{pos} = \Vert p_{a} - p_{b}\Vert_2$, $e_{\text{tilt}} = \cos^{-1}(\mathbf{n}_\text{a}^\top \mathbf{n}_\text{b})$, and $e_{\text{spin}} = \cos^{-1}\big({(\text{proj}_{{\pi}_\text{thorax}}\mathbf{x}_{a})}^\top  {(\text{proj}_{{\pi}_\text{thorax}}\mathbf{\mathbf{x}}_{b})}\big)$.

\begin{table}[tbh]
  \centering
  \caption{Error distribution reported as Mean $\pm$ Standard Deviation. Units: $e_\text{pos}$ in mm,  $e_\text{orient}$ in degee}
  \label{tab:error_stats}
  \begin{tabular}{lccc}
    \toprule
    \textbf{Metrics} &
    \textbf{${e}_{depth}$} &
    \textbf{${e}_{anatomy}$} &
    \textbf{${e}_{utility}$} \\
    \midrule
    Supine position $e_\text{pos}$                & $14.4 \pm 10.6$ & $42.5 \pm 13.0$ & $43.9 \pm 15.1$ \\
    Left lateral decubitus $e_\text{pos}$         & $15.8 \pm 8.0$ & $39.0 \pm 15.3$ & $36.1 \pm 19.3$ \\
    Subject-level mean $e_\text{pos}$             & $15.1 \pm 3.1$ & $41.1 \pm 9.1$ & $40.3 \pm 11.4$ \\
    All positional errors $e_\text{pos}$          & $15.0 \pm 9.5$ & $41.0 \pm 14.1$ & $40.7 \pm 17.4$ \\
    All Tilt errors $e_\text{tilt}$               & -- & $8.79 \pm 6.20$ & -- \\
    All Spin errors $e_\text{spin}$               & -- & $5.65 \pm 4.19$ & -- \\
    \bottomrule
  \end{tabular}
\end{table}

\section{Results and Discussion}\label{sec:results}

\begin{figure*}[tbh]
  \centering
  \subfloat[(a)]
  {\includegraphics[width=0.49\linewidth]{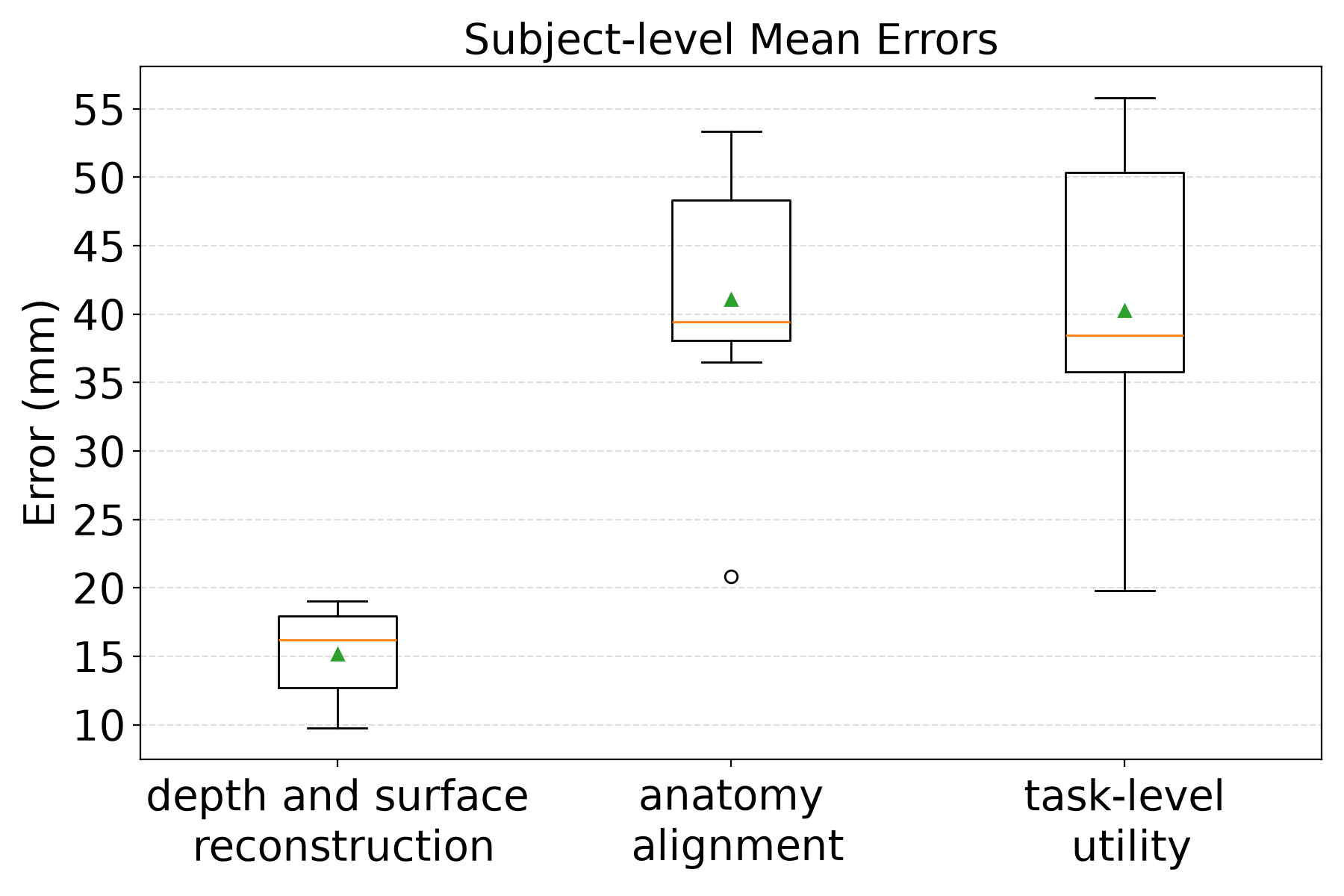}}
  \hfill
  \subfloat[(b)]
  {\includegraphics[width=0.49\linewidth]{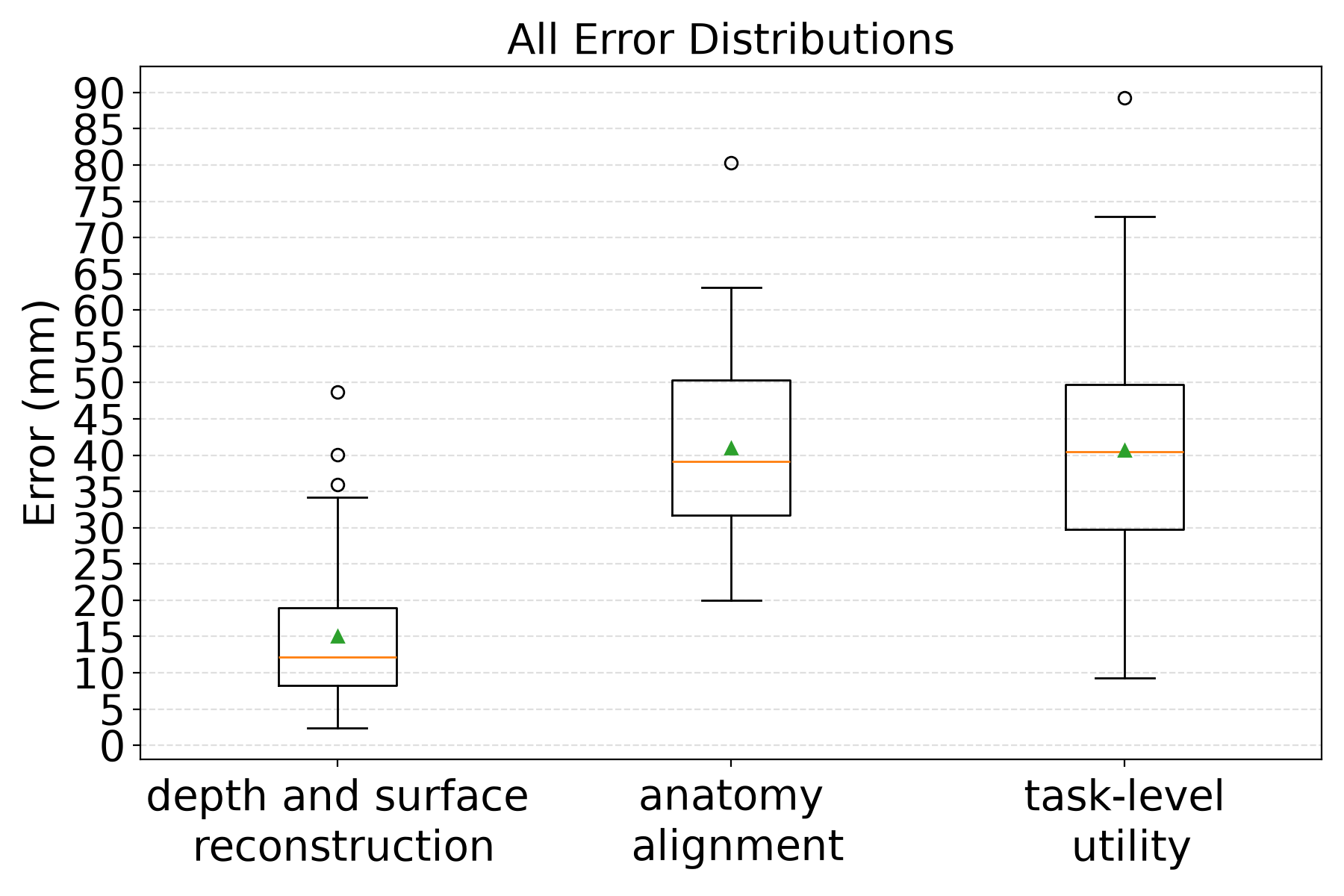}}
  \caption{(a) Error distribution per subject, (b) Error distribution of all probe poses. Detailed statistics are summarized in Table~\ref{tab:error_stats}.}
  \label{fig:boxplot_posture}
\end{figure*}

Figures~\ref{fig:boxplot_posture} and Table~\ref{tab:error_stats} show consistent trends across postures and subjects.
In all samples across supine and left lateral decubitus positions, the mean depth and surface errors are 15 mm, indicating that the predicted body surface closely approximates the patient’s actual body contour and that novice operators can reliably follow the MR overlay while maintaining probe–skin contact. The subject-level mean values demonstrate that average scan performance is consistent across examinations.
The mean anatomical positional error is approximately 40 mm, with substantial inter-subject variability in the initial placement of the A4C view. This finding highlights the persistent difficulty in robustly localizing the intercostal region. It should be noted that the accuracy of identifying the point of maximal impulse (PMI) by palpation at the fifth intercostal space remains dependent on operator experience, and the A4C view can be acquired from either the fourth or fifth intercostal space. Consequently, greater variability in the anatomical prediction is expected.
The mean tilt and spin orientation errors between the predicted and actual torso rotations were within 9° and 6°, respectively. These results suggest that patient registration provides an accurate estimate of torso pose, supporting its use for guiding the initial orientation of probe placement.
The utility error mostly follows the anatomical error means with a wider distribution, suggesting that anatomical alignment is the dominant source of error, with variation in user compliance with the guidance.

\textbf{Guidance Utility for Initial Placement.}
Beyond absolute error, we consider whether the proposed guidance reaches an accuracy range that is practically meaningful for novice setup.
Prior FoCUS training study reports that untrained physicians exhibit median probe placement errors of 3.2\, cm and 3.1\, cm after training with a simulator \cite{morgandoSimulatorBasedTrainingFoCUS2019}.
Our pilot results suggest that the proposed MR guidance can achieve placement errors proximal to the range, supporting the hypothesis that it can provide a clinically viable ``warm start'' prior to expert fine guidance. 
Future studies will quantify gains in consistency across novices and trials and evaluate downstream outcomes, such as time to first diagnostic view, displacement from the standard view location, and ultrasound image comparison between the predicted window and the standard view.

Our pilot findings suggest that anatomy-informed MR overlays can improve the setup phase of transthoracic teleoultrasound by providing a consistent initial probe placement on the patient surface, anchored to predicted thoracic intercostal landmarks.
This addresses a practical challenge in teleultrasound workflows, in which novice users struggle to identify initial windows, thereby delaying the expert’s examination.

\textbf{Limitations.}
We selected SAM 3D Body as the human mesh prior due to its stable and consistent performance. To mitigate residual depth ambiguities, we applied ICP registration with depth constraints and performed robust, body-part–wise point cloud optimization-based model fitting. The overall runtime is currently incurred in the model conversion and iterative optimization procedures. This overhead could be reduced by incorporating direct feedforward skeleton prediction, as suggested in \cite{liSKELCFCoarsetoFineBiomechanical2025}. In the present workflow, registration is intended as a one-time procedure during the operational setup and does not compensate for body motion occurring during the scan. Future work should therefore integrate motion tracking and motion compensation. Additionally, our pilot evaluation in healthy volunteers does not capture the full range of clinical variability (including gender diversity, body habitus, pathological conditions, and challenging acoustic windows), underscoring the need for larger, more heterogeneous clinical studies.

\subsection{Ablation Study}
To quantify the contribution of each component in the proposed registration pipeline, we conducted an ablation study over three configurations, evaluated with the depth-and-surface reconstruction error and the anatomy alignment error described above. The results are summarized in Table~\ref{tab:ablation_stats}.

\begin{table}[tbh]
  \centering
  \caption{Ablation study of the registration pipeline. Error distribution reported as Mean $\pm$ Standard Deviation.}
  \label{tab:ablation_stats}
  \begin{tabular}{lcc}
    \toprule
    \textbf{Metrics ${e}_{pos}$} &
    \textbf{Depth \& surface} &
    \textbf{Anatomy} \\
    \midrule
    RGB only (No ICP \& $\mathcal{P}_{\text{bparts}}$)            & 36.0 $\pm$ 17.6 & 51.0 $\pm$ 22.0 \\
    SAM3DB + ICP (No $\mathcal{P}_{\text{bparts}}$)                               & 22.1 $\pm$ 9.1  & 43.8 $\pm$ 16.3 \\
    \textbf{Full} (SAM3D + ICP + $\mathcal{P}_{\text{bparts}}$)        & \textbf{15.0} $\pm$ 9.5  & \textbf{41.0} $\pm$ 14.1 \\
    \bottomrule
  \end{tabular}
\end{table}

Fitting SMPL-X directly on the multiview SAM 3D Body predictions, which uses only RGB modalities, yields the largest errors, consistent with the systematic depth bias of monocular mesh recovery on HMD imagery. 
Adding depth-constrained ICP registration along the camera optical axis substantially reduces depth and surface errors and lowers anatomy error. This confirms that aligning the MHR meshes with the accumulated point cloud effectively corrects the absolute depth offset. However, without body-part–wise correspondence, the fitting remains sensitive to partial observations and local misalignment between the mesh and the point cloud, particularly for limbs and the torso boundary under in-bed postures. The full pipeline additionally incorporates the body-part matching loss, which achieves lower error with reduced variance, as part-wise correspondence constrains the fitting to semantically consistent matches.

Overall, the ablation indicates that the depth-constrained ICP contributes the largest single improvement in depth accuracy, while the body-part segmentation terms provide complementary refinements that improve robustness of the final SMPL-X fitting. The anatomy alignment error improves more modestly across configurations, suggesting that residual anatomical error is dominated by inter-subject landmark variability rather than surface registration accuracy, consistent with the discussion in Section~\ref{sec:results}.

\section{Conclusion}\label{sec:conclusion}

We presented an MR-based framework for automated patient registration and anatomy-informed initial probe placement guidance for transthoracic cardiac and lung teleultrasound using RGBD-based human mesh recovery and skeleton landmarks. 
We introduce a hybrid feedforward–optimization registration method that leverages SAM-based HMR predictions from multiview RGB observations as spatial priors and registers them to the measured point cloud, enabling robust and faithful patient registration.
By rendering the plane-specific probe pose guidance and body mesh {\em in situ}, the method provides a practical warm start for novice setup before expert takeover.
Although the system is designed to use an MR headset as the interface, paired with a freehand ultrasound probe, the proposed PIPG framework is intrinsically system-agnostic and could be adopted in most teleultrasound systems equipped with a calibrated RGBD camera (e.g., teleguided, teleoperated, or robotic ultrasound systems) to streamline setup and provide initial placement.
Pilot results with healthy volunteers motivate further clinical validation and downstream evaluation in teleoperated ultrasound.
Future work will evaluate clinically meaningful endpoints, such as time-to-first-diagnostic-view and image quality under expert supervision, and extend the guidance rules to additional scan planes and patient positions.
We will also explore personalized and semi-autonomous probe guidance, using the predicted surface and skeleton features as priors for locating patient-specific acoustic windows.







\bmsubsection*{Conflicts of Interest}

The authors declare no conflicts of interest.

\bibliography{miccai_2026_pipg}




\end{document}